\documentclass[conference]{IEEEtran}
\IEEEoverridecommandlockouts
\usepackage{cite}
\usepackage{amsmath,amssymb,amsfonts}
\usepackage{algorithmic}
\usepackage{graphicx}
\usepackage{textcomp}
\usepackage{array}
\usepackage[table]{xcolor}
\usepackage{cellspace}
\usepackage{booktabs}
\usepackage{balance}
\usepackage{xurl}
\usepackage{hyperref} 
\def\BibTeX{{\rm B\kern-.05em{\sc i\kern-.025em b}\kern-.08em
    T\kern-.1667em\lower.7ex\hbox{E}\kern-.125emX}}
\newcolumntype{P}[1]{>{\centering\arraybackslash}p{#1}}
\definecolor{lightgray}{gray}{0.9}
\begin{document}

\title{Public space security management using digital twin technologies\thanks{The research leading to these results has received funding from the European Union’s Horizon Europe research and innovation programme under grant agreement No 101073876 (Ceasefire). This publication reflects only the authors views. The European Union is not liable for any use that may be made of the information contained therein. The authors would like to acknowledge FlexSim Software Products, Inc. for providing a student license for the FlexSim software package.}}

\author{
    \IEEEauthorblockN{
        Stylianos Zindros\IEEEauthorrefmark{1}, 
        Christos Chronis\IEEEauthorrefmark{1}, 
        Panagiotis Radoglou-Grammatikis\IEEEauthorrefmark{2},
        Vasileios Argyriou\IEEEauthorrefmark{3}, 
    }
    
    \IEEEauthorblockN{
        Panagiotis Sarigiannidis\IEEEauthorrefmark{2}, 
        Iraklis Varlamis\IEEEauthorrefmark{1},
        Georgios Th. Papadopoulos\IEEEauthorrefmark{1}
    }
    
    \IEEEauthorblockA{
        \IEEEauthorrefmark{1}Department of Informatics and Telematics, Harokopio University of Athens, Athens, Greece\\
        \IEEEauthorrefmark{2}Department of Electrical and Computer Engineering, University of Western Macedonia, Kozani, Greece\\
        \IEEEauthorrefmark{3}Department of Networks and Digital Media, Kingston University, Kingston upon Thames, United Kingdom
    }

    \IEEEauthorblockA{
        Emails: itp23106@hua.gr, chronis@hua.gr, pradoglou@uowm.gr, vasileios.argyriou@kingston.ac.uk,\\
        psarigiannidis@uowm.gr, varlamis@hua.gr, g.th.papadopoulos@hua.gr
    }
}
\IEEEoverridecommandlockouts
\IEEEpubid{\makebox[\columnwidth]{978-1-5386-5541-2/18/\$31.00~\copyright2018 European Union \hfill} \hspace{\columnsep}\makebox[\columnwidth]{ }}

\maketitle

\IEEEpubidadjcol

\begin{abstract}
    As the security of public spaces remains a critical issue in today's world, Digital Twin technologies have emerged in recent years as a promising solution for detecting and predicting potential future threats. The applied methodology leverages a Digital Twin of a metro station in Athens, Greece, using the FlexSim simulation software. The model encompasses points of interest and passenger flows, and sets their corresponding parameters. These elements influence and allow the model to provide reasonable predictions on the security management of the station under various scenarios. Experimental tests are conducted with different configurations of surveillance cameras and optimizations of camera angles to evaluate the effectiveness of the space surveillance setup. The results show that the strategic positioning of surveillance cameras and the adjustment of their angles significantly improves the detection of suspicious behaviors and with the use of the DT it is possible to evaluate different scenarios and find the optimal camera setup for each case. In summary, this study highlights the value of Digital Twins in real-time simulation and data-driven security management. The proposed approach contributes to the ongoing development of smart security solutions for public spaces and provides an innovative framework for threat detection and prevention.
\end{abstract}

\begin{IEEEkeywords}
    Digital Twins, Public Space Security, Threat Scenarios, Surveillance Systems
\end{IEEEkeywords}

\section{Introduction} \label{intro}
    The security of public spaces is crucial in today's world, as criminal activity and terrorist attacks are on the rise worldwide \cite{b1}. Urban environments such as transportation hubs, shopping centers, and cultural institutions are essential components of social and economic life, but remain vulnerable to security threats \cite{b2}. Dealing with these risks requires innovative approaches that utilize technological advancements for enhanced surveillance and threat detection \cite{b3}, \cite{b4}.
    
    In recent years, Digital Twin (DT) technology has emerged as a promising solution for security management, providing real-time monitoring, analysis and predictive capabilities \cite{b5}. In the context of public space security, DTs facilitate data-driven insights through sensor integration, scenario simulation and predictive analytics \cite{b6}. The application of DTs in security management has attracted significant attention from both academia and industry. In addition, strategic collaborations have led to the development of DT-based security platforms in transportation infrastructure \cite{b7}, \cite{b8}. DTs have also been used for energy management, traffic monitoring and emergency response in smart cities \cite{b9}, as well as for visitor safety in shopping centers \cite{b10} and risk mitigation in industrial environments \cite{b11}, \cite{b12}.
    
    The ability of DTs to continuously update information, simulate risk scenarios and support real-time decisions makes them a valuable tool for crisis prevention and security management. Their proven effectiveness in various sectors underlines their potential to mitigate security threats in high-risk environments. This paper explores the role of DTs in improving security management in metro stations — critical public spaces characterized by high passenger density and increased vulnerability to security incidents \cite{b13}.
    
    The main objective of this research is the development of a Digital Twin of a metro station with a wide range of parameters that can support various security threatening scenarios. The DT model incorporates the station's infrastructure, passenger flows and surveillance camera data and is fully parametrized to allow several security threatening scenarios to be tested and help improve threat detection capabilities. By analyzing images captured by surveillance cameras, the model detects suspicious behavior and predict potential security threats. The aim is to improve security through real-time risk assessment and preventative measures, taking into account the challenges and limitations of DT implementation. This research utilizes the Digital Twin-as-a-Security-Service (DTaaSS) framework \cite{b6} to digitally represent the structure and operation of the corresponding metro station. The main contributions of this work include: \textbf{(i) Digital Representation:} A detailed DT of the M2 metro station ``Agios Ioannis” Athens, Greece, was developed in FlexSim, accurately modeling its layout and operations; \textbf{(ii) Simulation of Passenger Flow:} The model simulates passenger loads at different times of day, taking into account different camera placements and angles in order to assess their detection capabilities; \textbf{(iii) Evaluation of Security Performance}: Camera performance is analyzed based on density, angle and distance, while various security scenarios are simulated to identify potentially suspicious behavior.

    The remainder of the paper is organized as follows: Section \ref{lit_review} provides a detailed literature review on passenger flow management, DT technology and security applications. Section \ref{method} outlines the methodology, including the development and implementation of the Digital Twin model. Section \ref{results} presents the results and analysis, while Section \ref{conclusion} summarizes the key findings and discusses future research directions.

\section{Literature review} \label{lit_review}
    \subsection{Passenger Flow Management in Transportation Systems}
        Improving the security of transportation systems has become an important topic in the research community in recent years and is attracting continued interest \cite{b14}. With regard to the management of passenger flows in transport systems, simulation studies have been developed with the aim of optimizing the design to reduce congestion and overcrowding \cite{b15}, \cite{b16} and improving the management of pedestrian traffic in railway stations \cite{b17}. The development of smart railway stations is increasingly relying on Digital Twin technology to enhance passenger experience. This approach involves creating a virtual replica of the station that integrates real-time data, predictive models and various applications to optimize operations. Li et al. \cite{b18} propose a framework for the implementation of DTs in railway stations, describing the key data sets, the models used and the applications required to improve passenger flow and station management.
        
        Complementing this, Pokusaev et al. \cite{b19} propose a framework for discrete-event modeling of metro systems that simulates passenger boarding and alighting behavior and route selection to optimize planning and eliminate bottlenecks. Their model serves as a foundational component for DT implementations and enables dynamic adjustments to operating protocols under different demand scenarios. Extending these efforts, Padovano et al. \cite{b20} demonstrate a holistic DT platform for railway stations that integrates real-time data streams from surveillance systems, occupancy sensors and environmental monitors. By simulating emergencies (e.g. evacuations) and congestion at peak times, vulnerabilities are identified and preventative measures, such as rerouting crowds or staggered access protocols, are prescribed.

    \subsection{Digital Twin Technology}
        Digital Twins are increasingly being used to simulate, monitor and secure complex systems by integrating real-time data with advanced analytics. Their use in securing, Unmanned Aerial Vehicles (UAVs) and Industry 4.0 improves system resilience and facilitates infrastructure management.
        
        Eckhart and Ekelhart \cite{b21} emphasize the importance of embedding security measures into DTs from the initial design phase. Their study proposes an approach that integrates security mechanisms to protect data, ensure system integrity and address potential threats in virtual environments.
        
        In terms of UAV security, Fraser et al. \cite{b22} investigate how Digital Twin architectures improve intrusion detection by analyzing UAV flight data for anomalies. Their approach strengthens the defense of UAVs against modern cyber threats by enabling real-time threat detection.
        
        In the context of Industry 4.0, O’Connell et al. \cite{b23} shed light on the challenges of managing highly complex, interconnected production systems. Their study shows how Digital Twins improve security and interoperability by facilitating the integration and management of smart manufacturing technologies in real time.
        
        Recent research highlights the role of Digital Twins in security management by leveraging their ability to analyze sensor data, detect vulnerabilities and optimize processes. Their predictive capabilities support real-time decision making and risk mitigation, ultimately increasing both security and operational efficiency.

    \subsection{Security Management and Digital Twins}
        Digital Twins are increasingly recognized for their role in improving the design, development and operation of infrastructures, leading to better economic, social and environmental outcomes \cite{b24}. Their integration improves risk mitigation and infrastructure resilience, making systems more efficient.
        
        More recently, Digital Twins have been explored for disaster risk management. Lagap and Ghaffarian \cite{b25} propose a conceptual framework for the use of Digital Twins in real-time decision making, scenario simulation and crisis coordination. Homaei et al. \cite{b26} further emphasize their role in optimizing system design and performance while identifying potential cybersecurity risks that require further research.
    
        Beyond cybersecurity, Digital Twins contribute to smart city management by optimizing urban infrastructure, traffic control, energy efficiency and security. Sousa et al. \cite{b27} emphasize their role in integrating AI and IoT technologies to enhance resilience of cities. Khajavi et al. \cite{b5} focus on Digital Twins for building security and demonstrate their effectiveness in detecting fires and anomalies in real time using advanced sensors and 3D modeling.
        
\section{Methodology} \label{method}
    The proposed methodology is based on the DTaaSS architecture \cite{b6} to assess the degree of vulnerability and prove the effectiveness of Digital Twin technologies in public space security management. Specifically, through the deployment of a DT, the structure and operation of the metro station is replicated, creating a dynamic environment for security analysis. This system enables the continuous monitoring and analysis of surveillance camera images and facilitates the identification of suspicious individuals who could pose a threat. We assume that known suspects, indistinguishable from regular passengers, enter the metro station and carry out their activities. The system relies on advanced camera capabilities and computer vision algorithms to detect subtle anomalies in their behavior, providing a concrete basis for timely intervention.

    \subsection{Digital Twin Solution}
\begin{figure}[t]
        \centering
        \includegraphics[width=1\linewidth]{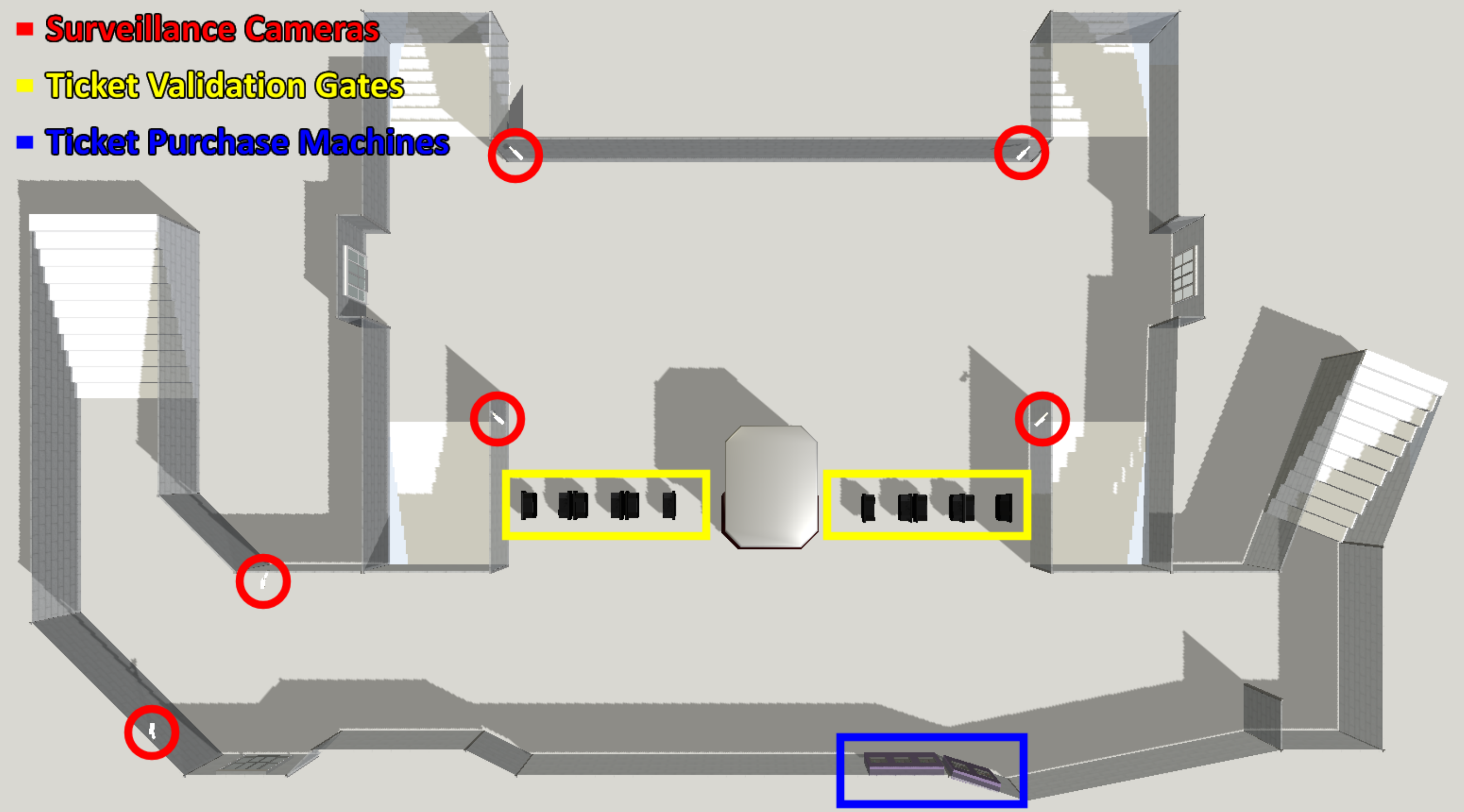}
        \caption{Digital Twin: Station’s Architectural Design}
        \label{fig:digitalTwin}
    \end{figure}

    The development begins with a comprehensive analysis of the station’s architectural design in order to create the DT using the FlexSim simulation software (Fig. \ref{fig:digitalTwin}). This initial assessment serves as the basis for the development of an accurate 3D model that replicates the spatial configuration of the station. The Base Model is equipped with six cameras that represents the current state of the metro station. Model 7 builds on this configuration and adds an additional camera. Model 9 further increases the number of cameras by integrating three additional cameras. Finally, Model 11 adds five more cameras to the setup compared to the Base Model.
    
    Following the creation of the models, passenger routes are simulated to examine movement patterns within the station, surveillance cameras are strategically placed at key observation points and delays at operational interfaces — such as ticket gates and ticket issuing machines — are simulated to assess their impact on the overall efficiency of the station.
    
    The detection process is driven by a probabilistic classifier based on the Detection Probability (P), which is defined using the parameters \( A_i^{norm} \), \( D_i^{norm} \) and \( N_i^{norm} \) as follows:
    \begin{equation}
      P_i = w_A \cdot A_i^{norm} + w_D \cdot D_i^{norm} + w_N \cdot N_i^{norm}
      \label{eq:detProb}
    \end{equation}
    where the weights \( w_D , w_A \) and \( w_N \) indicate how much each parameter contributes to the final \( P_i \) and satisfy:
    \begin{equation}
      w_A + w_D + w_N = 1
      \label{eq:weights}
    \end{equation}
    
    The \( A_i^{norm} \) represents the normalized angular deviation between the camera’s optical axis and the suspect’s movement direction. Since a smaller angular deviation should lead to a higher detection probability, we use an inverse normalization. The normalized distance \( D_i^{norm} \) denotes the distance between the suspect and the surveillance camera and the parameter \( N_i^{norm} \) corresponds to the normalized density of individuals within the camera’s field of view.
    \begin{align*}
         A_i^{\text{norm}} = 1 - \frac{A_i}{A_{max}}, && D_i^{\text{norm}} = \frac{D_i - 1}{D_{max}}, && N_i^{\text{norm}} = \frac{N_i}{N_{max}}
    \end{align*}
    where \( A_i \in [0^\circ, 90^\circ] \) and \( A_{max}=90^\circ \), \( D_i \in [0, 18] \) and the maximum operational distance of the camera is \( D_{max}=18 \) meters, and \( N_i \in [1, 18] \) and the maximum expected number of individuals in the field of view is \( N_{max}=18 \) to counteract potential occlusion effects in crowded environments.
    
    In this framework, the angular range is limited to \( 0^\circ - 90^\circ \) to allow for captures from either side of the individual \cite{b28}. The measured angles within this interval are normalized to the unit interval \( [0, 1] \), so that a frontal view (\( 0^\circ \)) is mapped to 1 and a profile view (\( 90^\circ \)) to 0. Any angle that exceeds \( 90^\circ \) is assigned the value 0, as it is assumed that it does not provide any additional information value.
    
    The trajectory of a suspect is considered detected if at least one camera registers a probability of detection that reaches or exceeds a predefined threshold value T.
    \begin{equation}
        \max_{i=1, \ldots, m} P_i \geq T
        \label{eq:detection}
    \end{equation}
    
    \subsection{Threat Simulation Scenarios}
        To further evaluate the security effectiveness of the Digital Twin, simulation scenarios are developed to analyze potential terrorist movements within the station. These scenarios assess detection performance under different conditions and examine the factors that influence detection accuracy. Three key aspects are taken into account: \textbf{(i) Effect of Time of Day:} Detection accuracy can vary depending on passenger density at different times of day; \textbf{(ii) Influence of the Suspect’s Route:} The accuracy of the system is evaluated based on the trajectory of a suspect within the station. Three different scenarios are used; \textbf{(iii) Impact of the Number of Cameras:} The impact of the number of active surveillance cameras on detection accuracy is investigated by varying the number of cameras.
        
        \begin{table}[b]
            \centering
            \caption{Passenger Traffic Distributions by Time of Day}
            \label{tab:traffic-distributions}
                \begin{tabular}{Sc Sc Sc}
                    \toprule
                    \textbf{Time of Day} & \textbf{Entrance} & \textbf{Exit} \\
                    \cmidrule(lr){1-1} \cmidrule(lr){2-2} \cmidrule(lr){3-3}
                    Morning (06:00-12:00) & $\mathcal{N}(7, 1.5^2)$ per min & $\mathcal{N}(5, 2^2)$ per 5 min \\
                    Midday (12:00-18:00) & $\mathcal{N}(5, 1.5^2)$ per min & $\mathcal{N}(7, 2^2)$ per 5 min \\
                    Afternoon (18:00-24:00) & $\mathcal{N}(3, 1.5^2)$ per min & $\mathcal{N}(9, 2^2)$ per 5 min \\
                    \bottomrule
                \end{tabular}
        \end{table}

        \begin{table*}[t]
            \centering
            \caption{Overall Model Accuracy by Time of Day and Suspect's Route}
            \label{tab:metrics}
                \begin{tabular}{l Sc Sc Sc Sc Sc Sc Sc}
                    \toprule
                    \textbf{Model}    & \textbf{Overall} & \multicolumn{3}{c}{\textbf{Time of Day}} & \multicolumn{3}{c}{\textbf{Suspect’s Route}} \\
                    \cmidrule(lr){3-5} \cmidrule(lr){6-8}
                             &         & \textbf{Morning} & \textbf{Midday} & \textbf{Afternoon} & \textbf{Scenario 1} & \textbf{Scenario 2} & \textbf{Scenario 3} \\
                    \midrule
                    \rowcolor{lightgray}
                    \textbf{Base Model} & 0.74    & 0.72    & 0.75   & 0.73   & 0.71    & 0.75    & 0.73 \\
                    \textbf{Model 7}    & 0.79    & 0.77    & 0.81   & 0.79   & 0.84    & 0.77    & 0.77 \\
                    \textbf{Model 9}    & 0.89    & 0.90    & 0.88   & 0.88   & 0.94    & 0.87    & 0.81 \\
                    \textbf{Model 11}   & 0.91    & 0.91    & 0.91   & 0.90   & 0.94    & 0.90    & 0.90 \\
                    \bottomrule
                \end{tabular}
        \end{table*}
        
\section{Experimental Results} \label{results}
    \subsection{Experimental Setup and Parameters Setting}        
        The simulation model is configured to run until it generates a total of 1000 suspicious samples, with each sample created based on a normal distribution \( \mathcal{N}(8, 2.3^2) \) that applies independently to each time period. Passenger traffic is modeled to reflect the dynamics of the real world. In the Morning, high volume of traffic flows from the entrances to the platforms, at Midday the movement to and from the platforms is balanced and in the Afternoon a high volume of flows from the platforms to the exits. All passenger flows follow a normal distribution, as presented in Table \ref{tab:traffic-distributions}. It is assumed that both passengers and suspects move at a standardized walking speed of 1.4m/s, which reflects the typical dynamics of pedestrians in cities \cite{b28}.

        The parameters \( A_i \) and \( D_i \) are modeled by \( \mathcal{N}(5.02, 1.49^2) \), while \( N_i \) is constrained between 0 and 18 persons. Cameras in the DT models are configured with a resolution of 4MP, a focal length of 6mm and a field of view of 50°, achieving a maximum effective range of 19m,\cite{b29}, with the shortest detection distance set at 1 meter due to the placement of the cameras at a height of 2.5 meters. The suspect's behavior is differentiated into three scenarios: \textbf{(i) Scenario 1:} The suspect enters the station, observes the surroundings and leaves; \textbf{(ii) Scenario 2:} The suspect enters the station, issues a ticket and then goes to the boarding platforms; \textbf{(iii) Scenario 3:} The suspect enters the station and goes directly to the platforms.

        The simulation also takes into account delays at transit points. The waiting time at the ticket validation gates is simulated with a \( \mathcal{N}(5, 2^2) \), with a minimum limit of 1s to ensure realistic conditions. The time spent at the ticket issuing machines also follows a \( \mathcal{N}(12, 2^2) \), with a minimum of 6s to capture variations in passenger familiarity with the process.
        
        After experimental testing, we chose a Threshold (T) of 0.45 to ensure that our accuracy is consistent with modern monitoring face recognition systems \cite{b30}, \cite{b31}. This value optimizes the sensitivity of the system and reliably detects genuine suspicious behavior.

    \subsection{Comparative Analysis}
        Table \ref{tab:metrics} summarizes the results obtained with the datasets of the different models developed for the different times of the day and for the suspect’s route. The Base Model achieves moderate detection accuracy, ranging from 0.72 to 0.75, indicating consistent but lower effectiveness across all time periods. The accuracy of Model 7 improves noticeably, with values increasing to 0.77–0.81. Model 9 and Model 11 achieve the highest detection accuracy and are consistently above 0.88 across all times of the day. Model 11 has the best overall performance, with minimal variation between morning, midday and afternoon, achieving a peak accuracy of 0.91. Furthermore, the minimal variation in Model 11 across different times suggests that this model is more robust to temporal variation.
        
        As for the detection accuracy of the different models for tracking a suspect’s route within the station in three different scenarios, the Base Model shows moderate accuracy in all scenarios with values from 0.71 and 0.75, indicating that the Base Model has difficulty in consistently adapting to different movement scenarios. There is a notable improvement when moving to Model 7, particularly in Scenario 1, where accuracy jumps to 0.84. However, the improvements are less pronounced in Scenarios 2 and 3, suggesting that Model 7 may be more suited to specific movement patterns rather than generalizing well across all conditions. Model 9 clearly outperforms its predecessors, achieving 0.94 in Scenario 1 and 0.87 and 0.81 in Scenarios 2 and 3, respectively. In addition, Model 11 shows the most consistent and best performing results. It achieves 0.94 in Scenario 1 and a high accuracy of 0.90 in Scenario 2 and 3, indicating that Model 11 is more robust and adaptable to different suspect movement patterns.
        
        The overall performance of the developed models in detecting potential threats shows a clear trend towards higher detection accuracy with improved models. The Base Model provides a basic approach to detecting suspects with an accuracy of 0.74, but is not robust enough for demanding security applications. Model 7 uses more cameras improving accuracy to 0.79, but is still below the 85-90\% level common in high-precision surveillance systems \cite{b30}, \cite{b31}. With a detection accuracy of 0.89, Model 9 is in a range more comparable to previous studies \cite{b30}, \cite{b31}. Model 11 achieves a detection accuracy of 0.91, which is quite competitive with the most modern monitoring systems \cite{b30}, \cite{b31}.
        
\section{Conclusions} \label{conclusion}
    This paper highlights the effectiveness of Digital Twin technologies in improving public space security through advanced surveillance and prediction solutions. By simulating real-world conditions in a metro station, the proposed approach enables dynamic monitoring, threat detection and optimal security management strategies. The results show that the more advanced models are more responsive to temporal variations as they maintain high detection accuracy across different times of the day. Furthermore, Model 11 (i.e. extra 5 cameras) proves to be the most adaptive in tracking the movements of suspects and consistently outperforms the other models in various movement scenarios within the station. Compared to existing high-precision surveillance systems, the developed models, especially Model 11, exhibit a competitive level of accuracy, making them suitable for real-world security applications. These results underline the potential of a Digital Twin solution for improving security in public space security by enabling robust real-time monitoring and threat detection.

\balance


\begin{thebibliography}{00}
    \bibitem{b1} Python, A., Bender, A., Nandi, A. K., Hancock, P. A., Arambepola, R., Brandsch, J., \& Lucas, T. C. (2021). Predicting non-state terrorism worldwide. Science advances, 7(31), eabg4778. doi: 10.1126/sciadv.abg 4778
    \bibitem{b2} J. Coaffee, B. Moritz, A. V. Nevez, S. Ilum, N. Gebbeken, A. Schroder, M. Schluter, P. Warnstedt, M. Stewart et al. (2022), Security by Design: Protection of public spaces from terrorist attacks. EC: European Commission.
    \bibitem{b3} Mademlis, I., Mancuso, M., Paternoster, C., Evangelatos, S., Finlay, E., Hughes, J., ... \& Papadopoulos, G. T. (2024). The invisible arms race: digital trends in illicit goods trafficking and AI-enabled responses. IEEE Transactions on Technology and Society.
    \bibitem{b4} Konstantakos, S., Cani, J., Mademlis, I., Chalkiadaki, D. I., Asano, Y. M., Gavves, E., \& Papadopoulos, G. T. (2025). Self-supervised visual learning in the low-data regime: a comparative evaluation. Neurocomputing, 620, 129199.
    Khajavi, S. H., Tetik, M., Liu, Z., Korhonen, P., \& Holmström, J. (2023). Digital twin for safety and security: Perspectives on building lifecycle. IEEE Access, 11, 52339-52356.
    \bibitem{b5} Khajavi, S. H., Tetik, M., Liu, Z., Korhonen, P., \& Holmström, J. (2023). Digital twin for safety and security: Perspectives on building lifecycle. IEEE Access, 11, 52339-52356.
    \bibitem{b6} Stefanidou, A., Cani, J., Papadopoulos, T., Radoglou-Grammatikis, P., Sarigiannidis, P., Varlamis, I., \& Papadopoulos, G. T. (2024, December). Leveraging digital twin technologies for public space protection and vulnerability assessment. In 2024 IEEE International Conference on Big Data (BigData) (pp. 2836-2845). IEEE.
    \bibitem{b7} KONE (2024), Developing digital twin with rockstar mindset. \href{https://www.kone.com/en/news-and-insights/stories/developing-digital-twin-with-rockstar-mindset.aspx}{https://www.kone.com/en/news-and-insights/stories/developing-digital-twin-with-rockstar-mindset.aspx}
    \bibitem{b8} Airport Council International (2024), GMR Airports Unveils AI Powered Digital Twin to Transform Passenger Experience and Enhance Airport Operations. \href{https://www.aci-asiapac.aero/media-centre/news/gmr-airports-unveils-ai-powered-digital-twin-to-transform-p%20assenger-experience-and-enhance-airport-operations}{https://www.aci-asiapac.aero/media-centre/news/gmr-airports-unveils-ai-powered-digital-twin-to-transform-p20assenger-experience-and-enhance-airport-operations}
    \bibitem{b9} White, G., Zink, A., Codecá, L., \& Clarke, S. (2021). A digital twin smart city for citizen feedback. Cities, 110, 103064.
    \bibitem{b10} Vohra, M. (2022). Overview of digital twin. Digital Twin Technology: Fundamentals and Applications, 1-18.
    \bibitem{b11} Fang, X., Wang, H., Liu, G., Tian, X., Ding, G., \& Zhang, H. (2022). Industry application of digital twin: From concept to implementation. The International Journal of Advanced Manufacturing Technology, 121(7), 4289-4312.
    \bibitem{b12} Papadopoulos, G. T., Leonidis, A., Antona, M., \& Stephanidis, C. (2022, June). User profile-driven large-scale multi-agent learning from demonstration in federated human-robot collaborative environments. In International Conference on Human-Computer Interaction (pp. 548-563). Cham: Springer International Publishing.
    \bibitem{b13} Shvetsov, A. V., \& Shvetsova, S. V. (2017). Research of a problem of terrorist attacks in the metro (subway, U-Bahn, underground, MRT, rapid transit, metrorail). European Journal for Security Research, 2, 131-145.
    \bibitem{b14} Yang, S., Li, T., Gong, X., Peng, B., \& Hu, J. (2020). A review on crowd simulation and modeling. Graphical Models, 111, 101081.
    \bibitem{b15} Tang, Y., Jiang, Y., Yang, H., \& Nielsen, O. A. (2020). Modeling and optimizing a fare incentive strategy to manage queuing and crowding in mass transit systems. Transportation Research Part B: Methodological, 138, 247-267.
    \bibitem{b16} Anagnostopoulou, A., Tolikas, D., Spyrou, E., Akac, A., \& Kappatos, V. (2024). The analysis and AI simulation of passenger flows in an airport terminal: A decision-making tool. Sustainability, 16(3), 1346.
    \bibitem{b17} Dubroca-Voisin, M., Kabalan, B., \& Leurent, F. (2019). On pedestrian traffic management in railway stations: simulation needs and model assessment. Transportation research procedia, 37, 3-10.
    \bibitem{b18} Li, D., Yang, X., \& Xu, X. (2020). A framework of smart railway passenger station based on digital twin. In CICTP 2020 (pp. 2623-2634).
    \bibitem{b19} Pokusaev, O., Chekmarev, A., \& Namiot, D. (2021). On digital twin for metro system. In 2021 IEEE East-West Design \& Test Symposium (EWDTS) (pp. 1-5). IEEE.
    \bibitem{b20} Padovano, A., Longo, F., Manca, L., \& Grugni, R. (2024). Improving safety management in railway stations through a simulation-based digital twin approach. Computers \& Industrial Engineering, 187, 109839.
    \bibitem{b21} Eckhart, M., \& Ekelhart, A. (2018). Towards security-aware virtual environments for digital twins. In Proceedings of the 4th ACM workshop on cyber-physical system security (pp. 61-72).
    \bibitem{b22} Fraser, B., Al-Rubaye, S., Aslam, S., \& Tsourdos, A. (2021). Enhancing the security of unmanned aerial systems using digital-twin technology and intrusion detection. In 2021 IEEE/AIAA 40th Digital Avionics Systems Conference (DASC) (pp. 1-10). IEEE.
    \bibitem{b23} O’Connell, E., O’Brien, W., Bhattacharya, M., Moore, D., \& Penica, M. (2023, May). Digital twins: Enabling interoperability in smart manufacturing networks. In Telecom (Vol. 4, No. 2, pp. 265-278). MDPI.
    \bibitem{b24} Sargiotis, D. (2024). Harnessing Digital Twins in Construction: A Comprehensive Review of Current Practices, Benefits, and Future Prospects. Benefits, and Future Prospects (July 02, 2024).
    \bibitem{b25} Lagap, U., \& Ghaffarian, S. (2024). Digital post-disaster risk management twinning: a review and improved conceptual framework. International Journal of Disaster Risk Reduction, 104629.
    \bibitem{b26} Homaei, M., Mogollón-Gutiérrez, Ó., Sancho, J. C., Ávila, M., \& Caro, A. (2024). A review of digital twins and their application in cybersecurity based on artificial intelligence. Artificial Intelligence Review, 57(8), 201.
    \bibitem{b27} Sousa, B., Arieiro, M., Pereira, V., Correia, J., Lourenço, N., \& Cruz, T. (2021). Elegant: Security of critical infrastructures with digital twins. IEEE Access, 9, 107574-107588.
    \bibitem{b28} Giannoulaki, M., \& Christoforou, Z. (2024). Pedestrian Walking Speed Analysis: A Systematic Review. Sustainability, 16(11), 4813.
    \bibitem{b29} Li, X., Liu, J., Baron, J., Luu, K., \& Patterson, E. (2021). Evaluating effects of focal length and viewing angle in a comparison of recent face landmark and alignment methods. EURASIP Journal on Image and Video Processing, 2021, 1-18.
    \bibitem{b30} Paul, K. C., \& Aslan, S. (2021). An improved real-time face recognition system at low resolution based on local binary pattern histogram algorithm and CLAHE. arXiv preprint arXiv:2104.07234.
    \bibitem{b31} Kim, H. B., Choi, N., Kwon, H. J., \& Kim, H. (2023). Surveillance system for real-time high-precision recognition of criminal faces from wild videos. IEEE Access, 11, 56066-56082.
\end{thebibliography}
\end{document}